\documentclass[journal,article,accept,pdftex,moreauthors]{Definitions/mdpi}

\firstpage{1}
\pubvolume{1}
\issuenum{1}
\articlenumber{0}
\pubyear{2026}
\copyrightyear{2026}
\datereceived{ }
\daterevised{ }
\dateaccepted{ }
\datepublished{ }

\usepackage{amsmath}

\newcommand{\Lce}{\mathcal{L}_{\mathrm{CE}}}
\newcommand{\Lalpha}{\mathcal{L}_{\mathrm{ALPHA}}}
\newcommand{\qwen}[1]{Qwen2.5-Coder-#1B}

\Title{From Evaluation to Optimisation: Hierarchy-Aware Training Signals for CWE Prediction in Python}

\Author{Muntasir Adnan$^{1,}$*\orcidA{} Manile Srun$^{1}$\orcidC{} and Carlos C. N. Kuhn$^{1}$\orcidB{}}

\AuthorNames{Muntasir Adnan and Manile Srun and Carlos C. N. Kuhn}

\address{%
$^{1}$ \quad Open Source Institute, Faculty of Science and Technology, University of Canberra, Canberra, Australia.}

\corres{Correspondence: Adnan.adnan@canberra.edu.au}

\abstract{The original ALPHA benchmark~\cite{alpha} introduced a taxonomy-aware penalty for evaluating CWE-level vulnerability prediction in Python and proposed that the penalty could theoretically also serve as a training signal.
This paper provides that validation. 
We compare three delivery mechanisms — supervised fine-tuning, a dual-head classification loss, and reinforcement learning with a dense reward derived from the normalised penalty.
We find that supervised approaches consistently regress below the zero-shot baseline under distribution shift, while GRPO succeeds. 
Our best policy reduces the cumulative ALPHA penalty of Qwen2.5-Coder-7B on Security Hardening and Adversarial Testing (SVEN) dataset by 27.9\% under greedy decoding, and by 25.5\% under sampled decoding($p = 0.005$, Welch's $t$-test), reaching statistical parity with its $4.5\times$ larger zero-shot teacher.
We conclude that the value of a hierarchical penalty as a training signal depends less on the penalty itself than on the directness of its delivery.}


\begin{document}


\section{Introduction}
\label{sec: intro}
Large language models(LLM) are now widely used to generate code~\cite{islam-etal-2024-mapcoder, ddi, pycapsule}, and a growing body of evidence indicates that the code they produce frequently contains exploitable vulnerabilities~\cite{asleep, how_secure}. 
The prior ALPHA work \cite{alpha} argued that binary vulnerability detection -- labelling a function as vulnerable or not -- provides insufficient specificity for the iterative repair pipelines increasingly used to improve generated code~\cite{pycapsule, islam-etal-2024-mapcoder}, since effective repair requires specific feedback~\cite{llm_correct}. 
ALPHA~\cite{alpha} therefore evaluates predictions at the level of the Common Weakness Enumeration (CWE) and introduces a taxonomy-aware penalty under which a prediction that falls near the true CWE in the MITRE hierarchy is penalised less than one that falls in an unrelated region of the taxonomy.
Because this penalty is continuous and direction-aware, it is a natural candidate for a training objective rather than only an evaluation metric.
Furthermore, the penalty's direction-aware asymmetry explicitly encodes downstream utility by penalising the loss of diagnostic information (over-generalisation) more heavily than unwarranted specificity (over-specification).

While the original ALPHA paper theorised this penalty could function as a loss function,
the empirical viability of this approach was unexplored. 
Crucially, an effective evaluation metric does not automatically translate into an effective loss function. 
Even if a signal accurately measures a target, the optimisation mechanism can distort or discard that information before it reaches the model parameters. 
Therefore, realising the theoretical promise of this penalty depends substantially on how it is architecturally delivered during training. 
We explore three delivery mechanisms, which differ in whether the penalty enters the training loss and in whether the output vocabulary is restricted to classes observed during training ---
\begin{itemize}
    \item \textbf{Supervised fine-tuning (SFT)}: The penalty does not enter the loss; the model imitates the reasoning and CWE labels produced by a stronger teacher model, and the penalty is used only to score the result.

    \item \textbf{A dual-head supervised loss}: This is the architecture proposed in the original ALPHA paper. 
    A classification head is trained with a differentiable expected-penalty term, $\Lalpha$, added to cross-entropy (CE). 
    The penalty enters the loss, but the head is restricted to the CWE vocabulary observed in training.

    \item \textbf{Reinforcement learning (RL)}: The $\Lalpha$ penalty defines the reward which is 
    one minus the normalised penalty; and the output vocabulary is unrestricted.
\end{itemize}

To evaluate these delivery mechanisms, we train on the hierarchically diverse SecurityEval~\cite{siddiq2022seceval} and CVEfixes~\cite{cvefix} dataset, and evaluate on Security Hardening and Adversarial Testing (SVEN)~\cite{sven} dataset (Section~\ref{sec: exp_setup}). 
On a matched in-distribution split (SecurityEval 70/30\% train-test split), the classification head is highly effective when trained with CE and mean pooling.
However, under the distribution shift to SVEN, both supervised mechanisms degrade significantly, regressing below the base model's (\qwen{7}) zero-shot performance.
In contrast, Group Relative Policy Optimisation (GRPO) is the only mechanism that demonstrates robustness under shift. 
Across every base-backbone configuration, GRPO successfully reduces the cumulative ALPHA penalty relative to the baseline. 
Notably, the optimal configuration achieves a 27.9\% reduction in the penalty under greedy decoding (25.5\% under sampled decoding; $p = 0.005$, Welch's $t$-test), reaching parity with a model 4.5$\times$ its size.

We make two contributions ---
\begin{itemize}
    \item Evidence that the ALPHA penalty is an effective training signal, and that delivery determines whether its benefit is realised. 
    Optimising the penalty improves CWE prediction under distribution shift; supplying the same penalty through supervised mechanisms does not. 

    \item A reinforcement-learning configuration that improves CWE prediction under distribution shift -- a base-backbone GRPO policy with a dense, penalty-derived reward that reduces the SVEN ALPHA penalty of Qwen2.5-Coder-7B by 27.9\% and outperforms its $4.5×\times$ larger variant, together with the data and hyperparameter analysis that characterise it.
\end{itemize}

\section{Background}
\label{sec: background}
LLMs now assist a substantial share of everyday software development, yet a recurring body of evidence shows that the code they produce is frequently insecure~\cite{llm_correct, how_secure, asleep}. 
This has prompted two broad responses: steering models towards secure code, and detecting vulnerabilities after generation so that they can be repaired. 
SVEN~\cite{sven} and SafeCoder~\cite{instruction_tuning} exemplifly the former, which raises the proportion of secure completions from 59.1\% to 92.3\% without degrading functional correctness.
For detection, comparative studies report that LLMs and static application security testing (SAST) tools occupy complementary regimes: SAST tools achieve low detection rates but relatively few false positives, whereas LLMs detect far more vulnerabilities at the cost of higher false-positive rates~\cite{zhou2024comparison, alpha}. 
Prompting-augmented frameworks such as DLAP~\cite{dlap} attempt to combine these strengths in a feedback loop setting. 

Most SAST-versus-LLM comparisons reduce the task to binary classification~\cite{zhou2024comparison, quan2025empirical}, which is too coarse to drive targeted repair~\cite{llm_correct}. 
Our prior work, ALPHA~\cite{alpha}, addressed this limitation by introducing the first function-level Python benchmark that scores predictions at CWE-level granularity for both LLMs and SAST tools. 
Rather than treating every misprediction as equally wrong, ALPHA exploits the hierarchical structure of the MITRE CWE taxonomy~\cite{cwe}. 
This taxonomy-aware perspective aligns with a broader line of hierarchy-aware learning, in which penalising errors by their semantic distance within a class hierarchy yields ``better mistakes" than a flat loss~\cite{better_mistake}. 

RL has become the standard mechanism for aligning LLMs with objectives that are not directly differentiable. 
Building on preference-based RL~\cite{human_preferences} and the RL-from-human-feedback pipeline popularised by InstructGPT~\cite{instructgpt}, most implementations optimise a reward using proximal policy optimisation~\cite{ppo}. 
GRPO~\cite{deepseekmath} simplifies this recipe by discarding the learned value network and instead estimating the baseline from the relative scores of a group of sampled outputs, thereby reducing the computational overhead.

Applying RL to code generation is now well established -- 
CodeRL~\cite{coderl}, PPOCoder~\cite{exec_based}, RLTF~\cite{rltf}, and StepCoder~\cite{stepcoder} all treat execution, unit-test, or compiler feedback as a reward to improve functional correctness.
Reward-driven approaches aimed specifically at security, however, remain scarce. 
The closest examples train models to repair vulnerabilities using PPO with syntactic and semantic similarity rewards~\cite{llm_vul_rep_rl, llm_vul_rep_rl_2}, arguing that ordinary cross-entropy under-weights the few security-critical lines relative to the bulk of functional code. 
Concurrent work has begun to pair online RL with a vulnerability reward model~\cite{onlinerl}. 
To our knowledge, none of these approaches uses a hierarchy-aware CWE penalty as an RL reward. 
This is the gap the present work addresses: we take the original ALPHA penalty and optimise it directly as a GRPO reward, yielding a principled, taxonomy-aware training signal for CWE-specific vulnerability detection.

\section{Methodology}
\label{sec: method}

\subsection{The Training-Corpus Problem and Label Withholding}
\label{subsec: corpus}
Function-level, human-annotated Python-CWE datasets are small -- SecurityEval~\cite{siddiq2022seceval} contains 121 samples across 69 CWE classes, and the Python subset of SVEN~\cite{sven} contains 342 samples across four base classes.
Larger CWE-labelled corpora exist~\cite{cvefix}, but trade reliability for scale~\cite{sec_vul_qual}.
Their labels are typically derived from static-analysis tools, whose coverage and precision are known to be poor~\cite{alpha, sec_vul_qual, vul_dataset}, making them unsuitable as a source of trustworthy supervision.
This scarcity of reliable data creates two distinct problems for training.
First, training on a classification objective alone with no LM signal at all risks eroding the pretrained backbone's generative competence, collapsing it into a narrow classifier.
Second, independently of data quantity, raw Python code itself provides a weak LM signal since short Python functions contain very little that a code-pretrained backbone has not already seen.

The natural remedy is to supplement the LM and classification head with reasoning and explanations generated by a teacher model that has access to the true label.
This, however, introduces label leakage into the classification head -- if the explanation names the CWE, the language head reads the answer directly from its input.
Our protocol resolves this through label withholding. 
The teacher (\qwen{32}) receives the code and the true CWE label, and generates a mechanistic explanation under an explicit constraint to omit any CWE identifier. 
This yields two corpora: an \emph{SFT corpus}, in which the teacher's output retains the CWE identifier;
and a \emph{classifier corpus}, in which the identifier is withheld so that the classification head predicts under genuine uncertainty.
Both corpora share the same schema; the gold label is used only in loss computation.

\subsection{Dual-Head Architecture and Supervised Objective}
\label{subsec: arch}

We adopt \qwen{7} as a shared backbone -- a decoder-only, open-weight model with a native causal LM head and a published zero-shot ALPHA baseline, enabling direct comparison. 
The backbone is adapted with LoRA~\cite{hu2022lora} adapters (Appendix~\ref{app: imp}), with the base transformer weights frozen throughout. 
Two heads read from this backbone: a causal LM head that generates a mechanistic vulnerability explanation, and a classification head that predicts a CWE from pooled hidden states. 
The heads are trained in two sequential stages, each optimising a single
objective.
We employ two-stage training.
In Stage 1, the LM head is trained with next-token CE on the teacher-explanation tokens only, yielding the SFT backbone; its role is to regularise the shared representation.
In Stage 2, we load the SFT adapter, apply a second LoRA adapter, and attach the classification head to the pooled last-layer hidden states. 
Only Stage 2's LoRA adapter and the head are updated, under the classification loss.
The head is a multi-layer perceptron (MLP) mapping the pooled representation to the
$n{=}69$ CWE classes observed in the training corpus, with three hidden layers of
width $2056$, $1024$ and $512$ before the output projection. 
The LoRA-adapted backbone supplies the features; this deeper head captures non-linear structure in the pooled embeddings. 
We evaluate mean and last-token pooling~\cite{ennadir2026pool} to form the
pooled representation.
The Stage 2 objective combines a standard classification loss with the hierarchy-aware $\Lalpha$ term --
\begin{equation}
\mathcal{L}_{total} = \mathcal{L}_{CE} + \lambda \cdot \mathcal{L}_{ALPHA},
\qquad
\mathcal{L}_{ALPHA}(\theta) = \sum_{v \in V_{obs}} p_\theta(v \mid x)\,
  \hat{P}(v, c_{true})
\end{equation}
where $\hat{P}(v,c_{true})=P(v,c_{true})/P_{\max}$ is the ALPHA graph penalty~\cite{alpha} normalised to $[0,1]$. 
Both terms are differentiable and read the same class distribution; $\Lalpha$ is an expected penalty weighting each candidate misclassification by its probability and its taxonomic severity. 
For $n=69$, $\Lce\in[0,\ln n]\approx[0,4.23]$ while normalised $\Lalpha\in[0, 1]$. 
Therefore, CE would dominate by $3$--$4\times$ at $\lambda{=}1$, so we set $\lambda{=}3$ to place the terms on a commensurate scale.

A property left unanalysed in the original ALPHA proposal is that $\Lalpha$ is insufficient to train the classifier on its own in the supervised regime. 
The two logit gradients (derived in Appendix~\ref{app: grad}) expose why:
\begin{equation}
\frac{\partial \mathcal{L}_{ALPHA}}{\partial z_k} = p_\theta(k) [\hat{P}(k) - \Lalpha], \qquad \frac{\partial \Lce}{\partial z_k} = p_\theta(k) - \mathbf{1}[k=c_{true}].
\end{equation}
CE is corrective through the indicator $\mathbf{1}[k{=}c_{true}]$.
It applies an upward push of magnitude $1-p_\theta(c_{true})$ on the true class, strongest exactly when the model is most wrong and unconditional on the training phase.
$\Lalpha$ on the other hand is confirmatory.
Since $\hat{P}(c_{true})=0$, its push on the true class is $p_\theta(c_{true})\cdot \mathcal{L}_{ALPHA}$, which vanishes if $p_\theta(c_{true}) \to 0$.
It goes silent precisely when correction is most needed and favours the true class only while that class already leads (Appendix~\ref{app: grad_2}).
It can extend a lead but not create one.
A confidently-wrong prediction becomes a practical trap under a finite training budget and therefore $\Lce$ is load-bearing.

This pathology is specific to the supervised expected-penalty setting. 
Under GRPO $\Lalpha$ penalty enters through sampled rollouts and normalised advantages rather than a belief-weighted expected-penalty gradient, so it retains a corrective signal even when the model is confidently wrong; this is why the ALPHA penalty can drive learning on its own under RL but must be paired with CE under SFT. 
The head places probability only over the $n=69$ observed classes, so $\Lalpha$ can only redistribute mass within that fixed simplex -- when the deployment distribution is dominated by a class seen only a handful of times in training, hierarchical shaping cannot compensate for a dimension the model has barely learned.

This analysis makes three predictions 
for SFT: First, when the shared representation is weak, and the model's distribution is already committed, $\Lalpha$ can only rescale existing mass.
So, toggling it on or off should leave the score unchanged. 
Second, when the representation is strong, its effect should be small and refining rather than decisive.
Third, whichever class leads at initialisation should be reinforced, so a biased starting point should collapse toward that class rather than away from it.

\subsection{Reinforcement Learning with an ALPHA Reward}
\label{subsec: rl}
SFT optimises against labelled targets, encouraging adaptation to the training distribution at the expense of the base model's broader capabilities~\cite{unleashing}, with sufficient adaptation potentially leading to overspecialisation~\cite{guiding_ai} and catastrophic forgetting~\cite{cat_forgetting}.
RL removes these constraints.
The model generates a free-form analysis, the predicted CWE is extracted from the text, and the reward is computed from the penalty.

We view this as a contextual bandit~\cite {sutton2018reinforcement} setting in which the entire response is treated as a single action with a terminal reward, thereby avoiding token-level credit assignment.
The state $s$ is the prompt, the action $a$ is the complete response, the policy $\pi_\theta(a|s)$ is the language model, factorised autoregressively.
The objective is --
\begin{equation}
    J(\theta) = \mathbb{E}_{x \sim D, \text{ }a \sim \pi_\theta(\cdot | x)} [R(a, c_{true})]
    \label{eq: j_reward}
\end{equation}
For a response $a$ whose extracted CWE is $\hat c$ --
\begin{equation}
    R(a, c_{true}) = 1 - \frac{P(\hat c, c_{true})}{P_{max}} \in [0, 1]
\end{equation}
We optimise $J(\theta)$ with GRPO~\cite{deepseekmath}, chosen for three reasons: it is suited to a single-GPU budget; it requires no separate value network, estimating the advantage baseline from a group of completions for the same prompt; and a Kullback-Leibler (KL) divergence ~\cite{kl} ($\mathbb{D}_\mathrm{KL}$) anchor to a frozen reference policy to keep training stable.
For each prompt, we sample $G$ completions ($G \in \{8, 16\}$), score each and form group-relative advantages $A$ skipping any group whose reward standard deviation falls below $10^{-6}$ --
\begin{equation}
    A_{i, t} = \frac{R_i - \mathrm{mean}(R_1, . . ., R_G)}{\mathrm{std}(R_1, . . ., R_G)}
\end{equation}

Although the optimisation problem is formulated as maximising $J(\theta)$ in~\eqref{eq: j_reward}, the implementation minimises its negation (Equation~\eqref{eq: reward_to_loss}), consistent with gradient-descent optimisers such as AdamW~\cite{adam} -- 
\begin{equation}
    \mathcal{L}_i = - (A_i \cdot \overline{\log \pi_\theta} (y_i | x) - \beta \cdot \mathbb{D}_\mathrm{KL}(\pi_\theta || \pi_{\mathrm{ref}}))
    \label{eq: reward_to_loss}
\end{equation}
The contrast with the supervised penalty is structural. 
The dual-head gradient scales with $p_\theta(k)$, attenuating updates to low-probability classes. 
The GRPO policy gradient scales with $\nabla_\theta \log \pi_\theta(y_i|x) = \nabla_\theta \pi_\theta / \pi_\theta$ -- the inverse dependence.
A correct action sampled at low probability yields a large update rather than a vanishing one.
This is a statement strictly about gradient scaling rather than a guarantee that reinforcement learning is unconditionally superior at escaping local optima.
Additionally, the KL anchor independently constrains policy movement (Section~\ref{subsec: abl_ref_pol}), and improvement still requires the sampled group to contain a better action.
Beyond constraining the policy and stabilising training, this KL penalty prevents degenerate solutions -- such as the model defaulting to the most frequent class regardless of input (e.g., CWE-89, which alone accounts for 54\% of SVEN samples). 

\subsection{Use of AI Tools in Manuscript Preparation}
Generative AI tools (Claude Sonnet 4.5 and Claude Sonnet 5, Anthropic) were used during manuscript preparation for two purposes: 
First, to assist in drafting and refining explanatory text, including identifying inconsistencies between reported claims and their supporting data. 
Second, to assist with debugging during implementation. 
All AI-assisted output was reviewed and verified by the authors, who take full responsibility for the accuracy and integrity of the analysis and content.

\section{Experimental Setup}
\label{sec: exp_setup}
\textbf{Dataset: }\emph{SecurityEval}~\cite{siddiq2022seceval} (121 Python functions, 69 CWEs) is the supervised training source; an 85/36 train/test split (seed 42) was carved out before any corpus generation, so the teacher never saw the held-out samples.
\emph{SVEN}~\cite{sven} (342 samples; CWE-022, CWE-078, CWE-079, CWE-089) is the out-of-distribution benchmark -- its four classes are likely common enough in the base model's pretraining corpora but barely overlap the training corpus used in this study. 
\emph{CVEfixes}~\cite{cvefix} (964 Python samples) augments the RL pool; its distribution is dominated by CWE-918, CWE-352 and CWE-444. 
Three RL corpus strategies are compared: SecurityEval only (121); \emph{combined} (CVEfixes filtered to CWEs shared with SecurityEval, capped at 31 per CWE; 499); and the same without the cap (584; the 23 retained CWEs average $\sim$25 samples each). 
The combined corpus $\mathcal{D}_{\text{combined}}$ retains only CVEfixes samples whose CWE also appears in SecurityEval -- $\mathcal{D}_{\text{combined}} = \{\, x \in \mathcal{D}_{\text{CVE}} : \text{cwe}(x) \in D_{\text{SVE}} \,\}$
so that the added data deepens coverage of the target classes rather than introducing unrelated ones.

\textbf{Policy and Training: }The policy and backbone is \qwen{7}~\cite{qwen25}; the teacher that generates the explanation corpora is \qwen{32}~\cite{qwen25}. 
All fine-tuning is parameter-efficient (PEFT)~\cite{peft} via LoRA \cite{hu2022lora}.
Unless stated otherwise, all reported scores represent cumulative ALPHA values. 
In selected instances, mean per-sample scores are reported and are labelled accordingly.
We report two complementary settings ---
\begin{itemize}
    \item \textbf{Out-of-distribution:} Training on SecurityEval or the combined RL data and evaluate on SVEN -- the deployment-realistic regime.

    \item \textbf{In-distribution:} A $70/30$ SecurityEval split, with both teacher corpora generated only from the training portion so that the held-out samples are unseen.
\end{itemize}
The in-distribution setting is diagnostic, not a target in its own right -- it isolates whether the dual-head loss behaves as the gradient analysis predicts.
Deployment is where a trained model meets inputs unlike its training distribution, so distribution shift is this work's actual concern. 
SFT and the dual-head loss are evaluated in both regimes because their in-distribution behaviour explains why they fail under shift (Section~\ref{sec: results}). 
GRPO is evaluated only under shift: it was adopted because SFT showed no distribution-shift gains despite adequate in-distribution performance, so an in-distribution GRPO run was not carried out.

\textbf{GRPO: }GRPO sweeps group size $G \in \{8, 16\}, \text{ KL regularisation coefficeint }\beta \in \{0.02, 0.04\}$ and 2-6 epochs.
Completions are sampled at temperature 0.8 (ollama default) during training.
Unless otherwise stated, GRPO checkpoints are evaluated under greedy decoding, matching typical deployment practice; Section~\ref{subsec: result_grpo} additionally reports a sampled-decoding evaluation of the best configuration for direct statistical comparison with the sampled-decoding baseline.
Continued training runs reduce the learning rate from $5 \times 10^{-6} \text{ to } 2 \times 10^{-6}$ and retain the original base model as the frozen KL reference so that the KL bound accumulates over epochs rather than relative to the most recent checkpoint.
All experiments used a single NVIDIA A100(40 GB) GPU.

\section{Results}
\label{sec: results}
\subsection{Supervised Results: In-Distribution and Under Shift}
\label{subsec: result_in_dist}

On the matched $70/30$ SecurityEval split, mean pooling substantially outperforms last-token pooling, and the dual-head marginally outperforms the SFT generative baseline. 
Table~\ref{tab: heldout} reports the held-out SecurityEval evaluation.
\begin{table}[H]
\caption{In-distribution evaluation on the held-out SecurityEval split. 
Lower is better. 
\label{tab: heldout}}
\begin{tabularx}{\textwidth}{lCC}
\toprule
\textbf{Condition} & \textbf{Cumulative ALPHA} $\downarrow$ & \textbf{Mean ALPHA (per sample)} \\
\midrule
Qwen2.5-Coder-7B-Instruct Baseline        & 230.7 & 6.41 \\
$\Lce$ only (last-token pooling)        & 233.6 & 6.49 \\
$\Lce$ only (mean pooling)              & 145.2 & 4.03 \\
$\Lce + \Lalpha$ (last-token pooling) & 233.6 & 6.49 \\
$\Lce + \Lalpha$ (mean pooling)       & \textbf{141.6} & \textbf{3.93} \\
\bottomrule
\end{tabularx}
\end{table}

Three observations follow: first, pooling is the dominant factor -- mean pooling reduces the penalty by roughly 40\% relative to last-token pooling.
Second, CE fine-tuning alone adds value over parsing the model's own free-form output when the distribution matches.
The mean-pooled classification head trained with only $\Lce$ reaches 145.2, improving on the zero-shot baseline of 230.7.
Third, adding $\Lalpha$ to the CE objective improves the mean-pooled score only marginally ($145.2 \rightarrow 141.6$), a difference of 0.1 per sample over 36 samples, which we do not claim as significant and on which we rest no conclusion.
This marginal improvement is the second prediction realised in Section~\ref{subsec: arch} -- once CE has fixed the ranking, the $\Lalpha$ term that only redistributes residual mass can do no more.
Under last-token pooling, switching the penalty term $\Lalpha$ on or off changes nothing (233.6 in both cases), consistent with the gradient analysis.
This is the first prediction of Section~\ref{subsec: arch} realised -- because the $\Lalpha$ gradient scales with $p_\theta(k)$, it can only amplify the distribution that the representation already produces; under a weak (last-token) representation, that distribution is fixed, and the penalty term has no distribution to reshape, and the score is identical.

Re-running the supervised mechanisms with evaluation on SVEN reverses the picture (Table~\ref{tab: dist_shift}). 
Both SFT and the dual-head classifier exceed the 7B zero-shot baseline's cumulative penalty of 753.2.

\begin{table}[H]
\caption{Out-of-distribution evaluation (SecurityEval $\rightarrow$ SVEN). 
Cumulative ALPHA, lower is better.
\label{tab: dist_shift}}
\begin{tabularx}{\textwidth}{lC}
\toprule
\textbf{Condition} & \textbf{Cumulative ALPHA} $\downarrow$ \\
\midrule
Qwen2.5-Coder-7B-Instruct Baseline          & \textbf{753.2} \\
$\Lce$ only (last-token pooling)            & 2123.5 \\
$\Lce$ only (mean pooling)                  & 3063.1 \\
$\Lce + \Lalpha$ (last-token pooling)       & 2312.7 \\
$\Lce + \Lalpha$ (mean pooling)             & 2799.2 \\
\bottomrule
\end{tabularx}
\end{table}

Two points are notable:
first, the ordering of the pooling strategies inverts relative to Table~\ref{tab: heldout} -- mean pooling, the strongest in-distribution, is the weakest under shift, because a representation tuned to the training distribution does not transfer when the test distribution is concentrated elsewhere. 
Second, under mean pooling, the effect of $\Lalpha$ is also larger under shift than in-distribution -- an 8.6\% reduction (3063.1 $\to$ 2799.2) -- though this does not reverse the classifier's overall degradation and the effect is not consistent across pooling strategies ($\Lalpha$ increases the penalty by 8.9\% under last-token pooling). 
The classifier conditions are the weakest overall -- it can predict only one of its 69 training classes, and SVEN's dominant class (CWE-89, 54\% of the test set) appeared twice in training.
This is the third prediction in Section~\ref{subsec: arch} realised -- with only two training instances of the dominant class, $p_\theta(\text{CWE-89})$ begins negligible, so the penalty gradient towards it is negligible and shrinks further as the model commits elsewhere.
The loss cannot lift a class the representation has not surfaced.

\subsection{GRPO under distribution shift}
\label{subsec: result_grpo}
Across all GRPO-initialised base-backbone configurations (6 independent, including configurations that overfit at later epochs) in Table~\ref{tab: full_rl}, every configuration improves on baseline; under a null hypothesis of no directional effect this outcome has probability $(0.5)^6 = 0.0156$.
The exceptions, which do not improve, are policies initialised from the SFT checkpoint and policies trained on unfiltered out-of-domain CVEfixes data; both are explained in Section~\ref{sec: abl}. 

The configuration that minimises the SVEN score most (base backbone; combined curated data; $G=16$, $\beta=0.02$; epoch 3) scores 543.1 under greedy decoding, a 27.9\% reduction relative to the 7B zero-shot baseline of 753.2, or 5.32 baseline standard deviations below the baseline mean.
Crucially, the conclusion that GRPO improves under shift does not depend on this selection -- at the default configuration with only SecurityEval training data, the improvement is already 8.3\% or 1.6 baseline standard deviations below the baseline mean, which establishes the direction of the effect on its own. 
Hyperparameter selection affects only the magnitude of the improvement, not whether it occurs.

Since the published baseline was evaluated under sampled decoding, we additionally evaluated this configuration under matched-sampled decoding (temperature 0.8) across six independent runs (526.0, 541.5, 548.2, 577.8, 586.4, 587.8; mean 561.3, SD 26.1), a 25.5\% reduction relative to baseline.
A Welch's two-sample $t$-test comparing this sampled distribution against the baseline's three published runs (mean 753.2, SD 39.5) gives $t = -7.62$ ($df = 2.9$, Welch--Satterthwaite), $p = 0.005$ (two-tailed); even the weakest of the six sampled runs (587.8) clears the baseline mean by 4.2 baseline standard deviations. 
We report 543.1 (greedy) as the deployment-realistic operating point and 561.3 (sampled, protocol-matched) as the figure supporting the statistical comparison. 
Both values lie within the $4.5\times$ larger 32B model's reported performance ($574.7 \pm 42.4$), so the GRPO-trained 7B model reaches parity with a model more than four times its size that it did not match before training.

\begin{table}[H]
\caption{All GRPO configurations on SVEN, reporting cumulative ALPHA. 
``Cont.'' marks continued training from the preceding checkpoint with the original frozen reference.
\textsuperscript{*} signifies the no-cap training corpus.}
\label{tab: full_rl}
\begin{tabularx}{\textwidth}{lCCCCCC}
\toprule
\textbf{Backbone} & $\boldsymbol{G}$ & $\boldsymbol{\beta}$ & \textbf{Epochs} & \textbf{Training Data} & \textbf{SVEN} $\downarrow$ & $\Delta$ (\%) \\
\midrule
7B Baseline & - & - & - & - & 753.2 & 0\% \\
SFT  & 8  & 0.04 & 2         & SecurityEval & 1450.5 & $-92.58\%$ \\
SFT  & 8  & 0.04 & 2         & Combined & 1377.8 & $-82.93\%$ \\
Base & 8  & 0.04 & 2         & CVEfixes & 1763.0 & $-134.07\%$ \\
Base & 8  & 0.04 & 2         & SecurityEval & 690.6 & $8.31\%$ \\
Base & 8  & 0.04 & 4 (cont.) & SecurityEval & 671.6 & $10.83\%$ \\
Base & 8  & 0.04 & 6 (cont.) & SecurityEval & 658.7 & $12.55\%$ \\
Base & 8  & 0.04 & 2         & Combined\textsuperscript{*} & 659.7 & $12.41\%$ \\
Base & 8  & 0.04 & 2         & Combined & 614.1 & $18.47\%$ \\
Base & 8  & 0.04 & 4 (cont.) & Combined & 609.7 & $19.05\%$ \\
Base & 8  & 0.04 & 6 (cont.) & Combined & 624.1 & $17.14\%$ \\
Base & 8  & 0.02 & 2         & Combined & 584.3 & $22.42\%$ \\
Base & 8  & 0.02 & 4 (cont.) & Combined & 571.4 & $24.14\%$ \\
Base & 16 & 0.04 & 2         & Combined & 570.1 & $24.31\%$ \\
Base & 16 & 0.04 & 4 (cont.) & Combined & 708.2 & $5.97\%$ \\
Base & 16 & 0.02 & 2         & Combined & 546.2 & $27.48\%$ \\
Base & 16 & 0.02 & 3 (cont.) & Combined & \textbf{543.1} & $\mathbf{27.89\%}$ \\
Base & 16 & 0.02 & 4 (cont.) & Combined & 642.1 & $14.75\%$ \\
\bottomrule
\end{tabularx}
\end{table}


\section{Ablations}
\label{sec: abl}
The GRPO study comprises many configurations; three factors determine the outcome: the reference policy, the composition of the training data, and the optimisation hyperparameters together with the number of epochs.

\subsection{Reference policy: base versus SFT backbone}
\label{subsec: abl_ref_pol}
The largest single effect is the choice of reference policy. 
Initialising GRPO from the SFT backbone ($\Lce$) records 1450.5 ALPHA penalty on SVEN; initialising from the base \qwen{7} records 690.6 on the same SecurityEval training data -- better by more than a factor of two. 
The cause is the KL anchor.
The SFT checkpoint is already biased towards CWE-89 (Section~\ref{subsec: rl}), and the KL term keeps the policy near that biased reference, so RL cannot escape the bias even with a correct reward. 
On the other hand, the base reference is not so biased and represents SVEN's four classes well from pretraining (based on baseline performance), so the KL term anchors the policy to a balanced starting point.

\subsection{Training data composition}
\label{subsec: abl_training_data}
SecurityEval provides only 121 training samples, so we tested whether enriching the RL training set with additional real-world Python vulnerabilities from CVEfixes improves the policy, and whether the composition of that enrichment matters.
Training with unfiltered CVEfixes does not help (1763.0, worse than SecurityEval alone) -- its dominant classes are absent from SVEN, so RL teaches the wrong distribution.
Restricting the enrichment to CVEfixes samples whose CWE also appears in SecurityEval (Section~\ref{sec: exp_setup}) improves the score.

This effect concerns the composition of the training data. 
It is distinct from the role of the KL term -- the KL anchor prevents the policy from collapsing onto a single high-frequency class to exploit the reward. 
Still, it does not, and is not intended to, counteract a genuine change in which classes the training examples reward. 
Enrichment and the KL term, therefore, act at different levels -- what the data emphasises, versus how far the policy may drift from the reference.

\subsection{Group size and KL coefficient}
\label{subsec: abl_g_size_kl}
Increasing the group size $G$ from 8 to 16 sharpens the advantage estimate, yielding more usable gradient updates per epoch. 
Lowering $\beta$ from 0.04 to 0.02 increases the reward's influence relative to the KL anchor. 
The two effects are approximately additive (Table~\ref{tab: hyperparam-ablation}).

\begin{table}[h]
\centering
\caption{Hyperparameter ablation (combined data, 2 epochs, base backbone).
The final column reports the change in cumulative ALPHA relative to the default GRPO configuration ($G{=}8,\beta{=}0.04$, first row); negative values indicate a lower penalty and hence better performance.}
\label{tab: hyperparam-ablation}
\begin{tabular}{cccc}
\toprule
$G$ & $\beta$ & SVEN $\downarrow$ & vs $G{=}8,\beta{=}0.04$ \\
\midrule
8  & 0.04 & 614.1 & 0 \\
8  & 0.02 & 584.3 & $-29.8$  \\
16 & 0.04 & 570.1 & $-44.0$  \\
16 & 0.02 & 546.2 & $-67.9$  \\
\bottomrule
\end{tabular}
\end{table}

Every configuration improves for several epochs and then degrades, as the policy starts to overfit.
We did not reserve a validation set for the reinforcement-learning runs: the peak epoch, like the other hyperparameters, was identified by its score on SVEN, which is the test benchmark. 
A validation set disjoint from the test data would allow principled early stopping without selecting on the test set; we identify this in Section~\ref{sec: limit_future}.

\section{Discussion}
\label{sec: discussion}
The organising result of this study is the contrast between Table~\ref{tab: heldout} and Table~\ref{tab: dist_shift}. 
The dual-head penalty loss is not incorrect -- it's the supervised mechanisms that lack robustness of delivery. 
SFT imitates a teacher whose labels and vocabulary carry their own biases; the dual-head loss delivers the penalty into a classifier whose vocabulary is fixed at training time. 
Both deliveries are mediated, and each mediator is a point of failure under shift.

GRPO is the only mechanism that improves under shift, and its advantage is consistent with two of its properties. 
Anchoring to a base reference allows the policy to reinforce knowledge already present from pretraining.
And unlike a binary or exact-match reward, which is uninformative when the model's prediction is wrong, the graded reward provides a useful gradient even for incorrect predictions, scaled by how taxonomically close the prediction is to the true label.

We have not isolated the contribution of the penalty's hierarchical structure within reinforcement learning, because we did not run a binary-reward GRPO baseline; it remains possible that part of the improvement is attributable to reinforcement learning with a free output vocabulary rather than to the graded penalty specifically. 
We therefore frame the contribution as a demonstration that the penalty can serve as an effective training signal when delivered directly, and that the delivery mechanism is the decisive variable, while leaving the marginal value of the penalty's hierarchical structure within RL as an open question and future work (Section~\ref{sec: limit_future}).

\section{Threats to Validity}
\label{sec: threat}
The best-observed GRPO value (543.1, greedy decoding) was selected using the test results across the data recipe, group size, KL coefficient, and epoch, and is therefore an optimistic estimate. 
We mitigate this with a configuration-robust value that uses no test-informed choices (690.6, $\Delta = 8.3\%$), the direction-independent sign test across six independent configurations ($p=0.0156$), and a decoding-matched statistical comparison against the baseline's own protocol ($p=0.005$, Welch's $t$-test; Section~\ref{subsec: result_grpo}). 
Additionally, because these figures rely on a single training run, their stability across different training seeds has not been tested. 
This is distinct from the decoding variance we already reported and is left for future work (Section~\ref{sec: limit_future}).

The in-distribution comparison of $\Lalpha$ and CE rests on 36 samples and a 0.1-per-sample difference; accordingly, we treat the two losses as comparable in-distribution and draw no conclusion about the difference.

Additionally, the classifier cannot predict unobserved classes and no nearest-ancestor fallback is implemented other than applying max penalty.

\section{Limitations and Future Work}
\label{sec: limit_future}
The most acute limitation is data sparsity: 121 training samples spread across 69 distinct CWEs (a mean of fewer than two samples per class, with a long tail of singletons), and as few as two samples for SVEN's most common class. 
This is the binding constraint on all three mechanisms and the principal motivation for future work -- larger labelled corpora, augmentation to balance CWE coverage, or an inference-time nearest-ancestor fallback for classes the classifier has not observed.

Furthermore, the evaluation is function-level Python on two datasets (SecurityEval, 121 samples; SVEN, 342 samples). 
The findings may not transfer to other languages or to repository or commit-level granularity. 
The auxiliary CVEfixes labels are inherited from CVE records and are noisier than the human-annotated evaluation labels.

Three further extensions follow directly from the analysis above: a binary-reward GRPO baseline to isolate the contribution of the graded penalty (Sections~\ref{sec: discussion}); a validation set disjoint from the test benchmark for principled early stopping, which would make the best-observed value selection-free (Section~\ref{subsec: abl_g_size_kl}); and reporting RL results across multiple seeds would establish the stability of the headline result (Section~\ref{sec: threat}). 
Beyond these, the longer-term direction is to deploy the trained policy within an iterative repair pipeline and to measure remediation success directly, which is the application that motivates CWE-level prediction in the first place.

\section{Conclusion}
This paper validated the proposal that the ALPHA penalty could serve as a training signal, and the validation succeeded in an unexpected way.
The proposed supervised mechanism fails: the expected-penalty loss leaves predictions essentially unchanged in-distribution, and the full supervised pipeline regresses below zero-shot under distribution shift for three diagnosable reasons: teacher-induced mode collapse, teacher gold label noise, and a fixed vocabulary that converts class sparsity into worst-case penalties, none of which implicates the penalty itself. 
Reinforcement learning with the penalty as a direct, dense reward eliminates all three mechanisms by construction and delivers an $8.3\%$ improvement under the default configuration, which used no test-informed choices and by $27.9\%$ under the best-observed configuration, bringing the 7B model to parity with one $4.5\times$ its size.
The ablations localise the gains: reference-policy choice dominates decisively, data curation comes second, group size and KL strength contribute near-additively, and early stopping is mandatory because training reward misleads.
The design principle for hierarchy-aware training that follows is therefore not that the penalty matters little, but that a useful penalty is squandered unless it is delivered directly to the model's parameters; for CWE prediction under realistic distribution shift, reinforcement learning provides that directness.

\authorcontributions{Conceptualization, M.A. and C.C.N.K.; methodology, M.A. and C.C.N.K.; software, M.A. and M.S.; validation, M.A. and M.S. and C.C.N.K.; formal analysis, M.A. and M.S.; investigation, M.A.; writing M.A and M.S. and C.C.N.K.; supervision, C.C.N.K. 
All authors have read and agreed to the published version of the manuscript.}

\funding{This work is funded under the agreement with the ACT Government, Future Jobs Fund -- Open Source Institute (OpenSI) -- R01553; and NetApp Technology Alliance Agreement with OpenSI -- R01657. 
This research was supported by the Australian Government through the Department of Education's National Industry PhD Program (project 36337). 
The views expressed herein are those of the authors and are not necessarily those of the Australian Government or the Department of Education.}



\dataavailability{Our code will be made available upon publication.}

\acknowledgments{During the preparation of this manuscript, the authors used Claude Sonnet 4.5 and 5 for debugging and refining the draft text. 
The authors have reviewed and edited all AI-assisted output and take full responsibility for the content of this publication.}

\conflictsofinterest{The authors declare no conflicts of interest.}

\appendixtitles{yes}
\appendixstart
\appendix

\section{Gradient Derivations for the Supervised Objective}
\label{app: grad}

This appendix derives the gradient of each loss term in Section~\ref{subsec: arch} with respect to the logit $z_k$ of class $k$.
These gradients govern how the classification head and the shared backbone are updated during training. 

\subsection*{Notation}
 
\begin{itemize}
    \item $z_k \in \mathbb{R}$: the raw logit for class $k$, produced by
          the linear classification head.
    \item $S = \sum_{j=1}^{n} e^{z_j}$: the softmax normalisation
          constant.
    \item $p_\theta(k) = \dfrac{e^{z_k}}{S}$: the model's predicted
          probability for class $k$ after softmax.
    \item $c_{\mathrm{true}}$: the ground-truth CWE class.
    \item $\hat{P}(v, c_{\mathrm{true}}) \in [0,1]$: the normalised ALPHA
          penalty for predicting class $v$ when the true class is
          $c_{\mathrm{true}}$; a constant with respect to model parameters.
    \item $\mathbf{1}[\cdot]$: the indicator function, equal to 1 if the
          condition holds and 0 otherwise.
\end{itemize}
 

\subsection{Softmax Derivative}
\label{app: softmax}
 
Both derivations rely on the derivative of the softmax output $p_\theta(v)$ with respect to logit $z_k$.
We derive this first using the quotient rule.
 
\paragraph{Case 1: $v = k$.}
 
\begin{align}
    \frac{\partial p_\theta(k)}{\partial z_k}
    &= \frac{\partial}{\partial z_k} \frac{e^{z_k}}{S} \notag \\[6pt]
    &= \frac{e^{z_k} \cdot S \;-\; e^{z_k} \cdot e^{z_k}}{S^2}
       \tag{quotient rule} \\[6pt]
    &= \frac{e^{z_k}}{S} - \frac{e^{z_k}}{S} \cdot \frac{e^{z_k}}{S}
       \notag \\[6pt]
    &= p_\theta(k) - p_\theta(k)^2 \notag \\[6pt]
    &= p_\theta(k)\bigl(1 - p_\theta(k)\bigr).
    \label{eq:softmax_same}
\end{align}
 
The numerator step uses $\frac{\partial}{\partial z_k} e^{z_k} = e^{z_k}$
and $\frac{\partial S}{\partial z_k} = e^{z_k}$ (since only the $k$-th
term in $S$ depends on $z_k$).
 
\paragraph{Case 2: $v \neq k$.}
 
\begin{align}
    \frac{\partial p_\theta(v)}{\partial z_k}
    &= \frac{\partial}{\partial z_k} \frac{e^{z_v}}{S} \notag \\[6pt]
    &= \frac{0 \cdot S \;-\; e^{z_v} \cdot e^{z_k}}{S^2}
       \tag{quotient rule; $\frac{\partial}{\partial z_k}e^{z_v}=0$} \\[6pt]
    &= -\frac{e^{z_v}}{S} \cdot \frac{e^{z_k}}{S} \notag \\[6pt]
    &= -p_\theta(v) \cdot p_\theta(k).
    \label{eq:softmax_diff}
\end{align}
 
\subsection{Gradient of \texorpdfstring{$\mathcal{L}_{\mathrm{CE}}$}{L\_CE}}
\label{app: ce_grad}

\begin{equation}
    \mathcal{L}_{\mathrm{CE}}
    = -\log p_\theta(c_{\mathrm{true}})
    = -\log \frac{e^{z_{c_{\mathrm{true}}}}}{S}
    = -z_{c_{\mathrm{true}}} + \log S \notag
    \label{eq:ce_def}
\end{equation}
 
 
\paragraph{Derivative of $\log S$ with respect to $z_k$.}
 
\begin{align}
    \frac{\partial \log S}{\partial z_k}
    &= \frac{\partial \log S}{\partial S} \cdot \frac{\partial S}{\partial z_k} \tag{chain rule}\\[6pt]
    &= \frac{1}{S} \cdot \frac{\partial S}{\partial z_k} \notag \\[6pt]
    &= \frac{e^{z_k}}{S}
       \tag{only the $k$-th term survives} \\[6pt]
    &= p_\theta(k) \text{ or } softmax(k). \notag
\end{align}
 
\paragraph{Case 1: $k \neq c_{\mathrm{true}}$.}
 
The term $-z_{c_{\mathrm{true}}}$ does not depend on $z_k$, so:
 
\begin{align}
    \frac{\partial \mathcal{L}_{\mathrm{CE}}}{\partial z_k}
    &= \frac{\partial}{\partial z_k}(-z_{c_{\mathrm{true}}})
     + \frac{\partial \log S}{\partial z_k} \notag \\[6pt]
    &= 0 + p_\theta(k) \notag \\[6pt]
    &= p_\theta(k). \notag
\end{align}
 
Since $p_\theta(k) > 0$, the gradient is positive and the logit is
pushed \emph{down}.
 
\paragraph{Case 2: $k = c_{\mathrm{true}}$.}
 
Now $-z_{c_{\mathrm{true}}}$ does depend on $z_k$, contributing $-1$:
 
\begin{align}
    \frac{\partial \mathcal{L}_{\mathrm{CE}}}{\partial z_k}
    &= \frac{\partial}{\partial z_k}(-z_{c_{\mathrm{true}}})
     + \frac{\partial \log S}{\partial z_k} \notag \\[6pt]
    &= -1 + p_\theta(k) \notag \\[6pt]
    &= p_\theta(k) - 1. \notag
    \label{eq:ce_correct}
\end{align}
 
Since $p_\theta(k) \in (0,1)$, the gradient is negative and the logit
is pushed \emph{up}.
 
\paragraph{Unified form.}
 
Both cases combine into:
 
\begin{equation}
    \frac{\partial \mathcal{L}_{\mathrm{CE}}}{\partial z_k}
    = p_\theta(k) - \mathbf{1}[k = c_{\mathrm{true}}].
    \label{eq: ce_grad}
\end{equation}
 
This gradient is always well-defined and is strongest early in training when $p_\theta(c_{\mathrm{true}})$ is small, providing the largest upward push to the correct logit precisely when the model is most uncertain.

\subsection{Gradient of \texorpdfstring{$\mathcal{L}_{\mathrm{ALPHA}}$}{L\_ALPHA}}
\label{app: alpha_grad}
 
\begin{equation}
    \mathcal{L}_{\mathrm{ALPHA}}(\theta)
    = \sum_{v \in V_{\mathrm{obs}}} p_\theta(v \mid x)\,\hat{P}(v, c_{\mathrm{true}}). \notag
    \label{eq:alpha_def}
\end{equation}
 
The penalties $\hat{P}(v, c_{\mathrm{true}})$ are constants with respect
to $\theta$; only $p_\theta(v)$ carries dependence on $z_k$.
 
\begin{align}
    \frac{\partial \mathcal{L}_{\mathrm{ALPHA}}}{\partial z_k}
    &= \sum_{v} \hat{P}(v, c_{\mathrm{true}})
       \cdot \frac{\partial p_\theta(v)}{\partial z_k}. \notag
\end{align}
 
Split the sum into the $v = k$ term and \textbf{all} $v \neq k$ terms, then substitute the softmax derivatives from
\eqref{eq:softmax_same} and \eqref{eq:softmax_diff}:
 
\begin{align}
    \frac{\partial \mathcal{L}_{\mathrm{ALPHA}}}{\partial z_k}
    &= \hat{P}(k, c_{\mathrm{true}}) \cdot p_\theta(k)\bigl(1-p_\theta(k)\bigr)
     + \sum_{v \neq k} \hat{P}(v, c_{\mathrm{true}})
       \cdot \bigl(-p_\theta(v)\,p_\theta(k)\bigr). \notag
\end{align}
 
Factor out $p_\theta(k)$:
 
\begin{align}
    &= p_\theta(k) \left[
         \hat{P}(k, c_{\mathrm{true}})\bigl(1 - p_\theta(k)\bigr)
         - \sum_{v \neq k} \hat{P}(v, c_{\mathrm{true}})\,p_\theta(v)
       \right]. \notag
\end{align}
 
Expand $\hat{P}(k, c_{\mathrm{true}})(1 - p_\theta(k))$:
 
\begin{align}
    &= p_\theta(k) \left[
         \hat{P}(k, c_{\mathrm{true}})
         -  \{ \hat{P}(k, c_{\mathrm{true}})\,p_\theta(k)
         + \sum_{v \neq k} \hat{P}(v, c_{\mathrm{true}})\,p_\theta(v) \}
       \right]. \notag
\end{align}
 
Observe that:
 
\begin{equation}
    \hat{P}(k, c_{\mathrm{true}})\,p_\theta(k)
    + \sum_{v \neq k} \hat{P}(v, c_{\mathrm{true}})\,p_\theta(v)
    = \sum_{v} \hat{P}(v, c_{\mathrm{true}})\,p_\theta(v)
    = \mathcal{L}_{\mathrm{ALPHA}} \notag
\end{equation}
 
\begin{align}
    \therefore \frac{\partial \mathcal{L}_{\mathrm{ALPHA}}}{\partial z_k}
    &= p_\theta(k) \left[
         \hat{P}(k, c_{\mathrm{true}}) - \mathcal{L}_{\mathrm{ALPHA}}
       \right].
       \label{eq: alpha_grad}
\end{align}
 
 
 
 

\section{Gradient Analysis of the Classification Objective}
\label{app: grad_2}

We take the $\Lce$ and $\Lalpha$ gradients from \eqref{eq: ce_grad} and \eqref{eq: alpha_grad}.
$\Lalpha$ is a state-dependent penalty.
$z_k$ is pushed up in \eqref{eq: alpha_grad} if class $k$'s penalty in $[\hat{P}(k, c_{\mathrm{true}}) - \Lalpha]$ lies below the current expected penalty $\Lalpha$. 
Additionally, the prefactor $p_\theta(k)$ scales this by the model's current belief in class $k$. 
For the true class, $\hat{P}(c_{true})=0$ gives
\begin{equation*}
    \frac{\partial \mathcal{L}_{ALPHA}}{\partial z_{c_{true}}} = -\,p_\theta(c_{true})\,\mathcal{L}_{ALPHA}
\end{equation*}
an upward push of magnitude $p_\theta(c_{true})\Lalpha$ that shrinks to zero if $p_\theta(c_{true})\to0$ which depends heavily on the training phase.

\subsection{When does ALPHA favour the true class?}
\label{app:crossover}
Let $k^\star$ be a close-but-wrong neighbour. 
The true class receives the stronger upward push when its gradient is more negative than the neighbour's.
So ideally:
\begin{align}
    -p_\theta(c_{\text{true}}) \Lalpha &< p_\theta(k^*) \left[ \hat{P}(k^*) - \Lalpha \right] \notag \\
\intertext{Multiplying both sides by $-1$ to get the magnitude flips the inequality and reverses the subtraction:}
    p_\theta(c_{\text{true}}) \Lalpha &> p_\theta(k^*) \left[ \Lalpha - \hat{P}(k^*) \right] \notag \\
\intertext{Dividing by $\Lalpha$:}
    p_\theta(c_{\text{true}}) &> p_\theta(k^*) \left( 1 - \frac{\hat{P}(k^*)}{\Lalpha} \right) \notag
\intertext{Since $k^*$ is a close neighbour, $\hat{P}(k^*)$ will be close to 0 and $\frac{\hat{P}(k^*)}{\Lalpha} \approx 0$, so:}
    p_\theta(c_{\text{true}}) &> p_\theta(k^*) \left( 1 - \frac{\hat{P}(k^*)}{\Lalpha} \right) \approx p_\theta(k^*)
    \label{eq: magnitude}
\end{align}

The contrast with CE is the entire point. 
CE's upward push on the true class is positive for every state due to the $\textbf{1}[\cdot]$ function; it is unconditional. 
ALPHA's push favours the true class only conditionally, when the state-dependent inequality \eqref{eq: magnitude} holds -- that is, only while the true class already leads its nearest wrong neighbour. 
A loss whose corrective effect is contingent on the model already being right cannot, by itself, make an incorrect model right; it can widen an existing margin but cannot open one.

\section{Implementation.}
\label{app: imp}
All phases train LoRA adapters via peft~\cite{peft} on a bfloat16~\cite{bfloat} \qwen{7} backbone transformer from the HuggingFace transformer library, with gradient checkpointing and manual device placement. 
The ALPHA penalty graph is taken from~\cite{alpha}; the teacher corpus is generated offline from \qwen{32} via the ollama client.
SFT and GRPO (Section~\ref{subsec: rl}) is implemented directly on \texttt{torch}.
SFT and GRPO hyperparameters are described in Table~\ref{tab: sft_hp} and Table~\ref{tab: rl_hp} respectively.

\begin{table}[h]
    \centering
    \caption{GRPO Hyperparameters}
    \begin{tabular}{ll}
    \toprule
    & Value \\
    \midrule
    Generation temperature & 0.8 \\
    Group size $G$ & $\{8, 16\}$ \\
    KL regularisation coefficient $\beta$ & $\{0.02, 0.04\}$ \\
    Epochs & 2-6 \\
    top\_p & 0.9 \\
    \bottomrule
    \end{tabular}
    \label{tab: rl_hp}
\end{table}

\begin{table}[!h]
\centering
\caption{SFT Hyper Parameters}
\label{tab: sft_hp}
\begin{tabular}{lll}
\toprule
 & SFT & Classifier \\
\midrule
Optimiser        & AdamW      & AdamW (trainable params only) \\
Learning rate    & $2\mathrm{e}{-4}$ & $1\mathrm{e}{-4}$ \\
Epochs           & 3          & 5 \\
Batch size       & 2          & 2 \\
Max length       & 1024       & 1024 \\
LoRA $r/\alpha/\text{drop}$ & $16/32/0.05$ & $16/32/0.05$ \\
LoRA Target Modules & q\_proj, k\_proj, v\_proj, o\_proj & q\_proj, k\_proj, v\_proj, o\_proj \\
Pooling          & ---        & mean $\mid$ last \\
$\lambda$        & ---        & 3.0 \\
Seed             & 42         & 42 \\
Precision        & bfloat16   & bfloat16 \\
\bottomrule
\end{tabular}
\end{table}

\reftitle{References}
\bibliography{reference}

\end{document}